\documentclass{article}
\usepackage{amsmath,graphicx, bm}
\usepackage{algorithm, algorithmic}

\usepackage{amsfonts} 
\usepackage{makecell} 

\usepackage{CJKutf8}
\usepackage{tikz}
\usetikzlibrary{arrows, calc}
\tikzstyle{int}=[draw, fill=blue!20, minimum size=2em]
\tikzstyle{init} = [pin edge={to-,thin,black}]

\PassOptionsToPackage{square}{natbib}

\usepackage[nonatbib,final]{nips_2018}

\usepackage[utf8]{inputenc} 
\usepackage[T1]{fontenc}    
\usepackage{hyperref}       
\usepackage{url}            
\usepackage{booktabs}       
\usepackage{amsfonts}       
\usepackage{nicefrac}       
\usepackage{microtype}      

\usepackage{subfigure}

\title{DONUT: CTC-based Query-by-Example Keyword Spotting}

\author{
  Loren Lugosch\\
  Fluent.ai\\
  \texttt{loren.lugosch@fluent.ai} \\
  \And
    Samuel Myer \\
  Fluent.ai\\
  \texttt{sam.myer@fluent.ai} \\
  \AND
    Vikrant Singh Tomar \\
  Fluent.ai\\
  \texttt{vikrant.tomar@fluent.ai} \\
}

\begin{document}

\maketitle

\begin{abstract}
Keyword spotting---or wakeword detection---is an essential feature for hands-free operation of modern voice-controlled devices. With such devices becoming ubiquitous, users might want to choose a personalized custom wakeword. In this work, we present DONUT, a CTC-based algorithm for online query-by-example keyword spotting that enables custom wakeword detection. The algorithm works by recording a small number of training examples from the user, generating a set of label sequence hypotheses from these training examples, and detecting the wakeword by aggregating the scores of all the hypotheses given a new audio recording. Our method combines the generalization and interpretability of CTC-based keyword spotting with the user-adaptation and convenience of a conventional query-by-example system. DONUT has low computational requirements and is well-suited for both learning and inference on embedded systems without requiring private user data to be uploaded to the cloud. 
\end{abstract}

\section{Introduction}
\label{sec:intro}
Keyword spotting enables a user to activate a device conveniently, without pushing a button or otherwise physically touching the device, by detecting when the user speaks a certain ``wakeword''. Most wakeword detectors use a pre-set wakeword (for example, ``Hey Siri'', ``OK Google'', or ``Alexa''). It is desirable for users to be able to choose their own wakeword instead of using a pre-set wakeword. For instance, if the device is a pet robot, a custom wakeword would allow users to give their robot a name, which gives the device a more personal feel. 

Neural networks can be trained to perform very accurate online keyword spotting for a pre-defined wakeword \cite{Chen2014, Myer}. However, to recognize a new wakeword, the network must be retrained, which requires time and many training examples. A more sample-efficient approach based on connectionist temporal classification (CTC) \cite{Graves2006} overcomes this problem by representing the keyword as a sequence of phonetic labels and using the CTC forward algorithm to efficiently compute a score for the sequence given the neural network output \cite{hwang2015online, Lengerich2016, Zhuang2016}. Thus, to recognize a custom wakeword, the user can provide the label sequence corresponding to the desired phrase without retraining the network.

A disadvantage of conventional CTC-based keyword spotting is that it is ``query-by-string'': that is, the desired wakeword must be provided in text form. Query-by-string keyword spotting is somewhat inconvenient for the user and requires that the device have a text interface in addition to the voice interface. 
Additionally, because the wakeword is provided through text, the wakeword model may not actually match the user's own pronunciation of the phrase.

A more natural method for custom wakeword detection is ``query-by-example'' keyword spotting. In a query-by-example system, the user teaches the system the desired wakeword by recording a few training examples, and the keyword spotter uses some form of template matching to compare incoming audios with these training examples to detect the wakeword. In dynamic time warping (DTW)-based keyword spotting, for example, a variable-length sequence of feature vectors, such as Mel-filterbank cepstral coefficients (MFCCs) \cite{snips} or phoneme posteriors \cite{PosteriorgramDTW,Zhang2009, Rodriguez-Fuentes2014}, is extracted from the query audio and test audio, and the DTW alignment score between query and test is used as the detection score. Other template-matching approaches compare fixed-length feature vectors, such as the final hidden states of a pre-trained recurrent neural network (RNN) \cite{Chen2015} or the output of a Siamese network \cite{Settle2016, Settle2017}, using the cosine distance.

Systems that use template matching are difficult to interpret, and therefore difficult to debug and optimize. For instance, it is hard to say why a keyword is incorrectly detected or not detected in a system based on dynamic time warping (DTW) simply by inspecting the DTW matrix. Likewise, the hidden states of RNNs can sometimes be interpreted (c.f. \cite{radford2017learning}, \cite{verwimp2018}), but this is currently only possible with some luck and ingenuity. In contrast, a CTC-based model is easy to interpret. The wakeword model itself is interpretable: it consists simply of a human-readable string, like ``\texttt{ALEXA}'' or ``\texttt{AH L EH K S AH}'', rather than a vector of real numbers. Inference is interpretable because the neural network outputs are peaky and sparse (the "blank" symbol has probability $\approx$1 at almost all timesteps), so it is easy to determine what the network ``hears'' for any given audio and whether it hears the labels of the wakeword \cite{bluche2015framewise}. This is a useful property because it enables the system designer to take corrective action. For instance, one might identify that a particular label is not well-recognized and augment the training data with examples of this label.

In this paper, we propose a new method for custom wakeword detection that combines the convenience and speaker-adaptive quality of query-by-example methods with the generalization power and interpretability of CTC-based keyword spotting. We call our method ``DONUT'', since detection requires $O(NUT)$ operations given the neural network output, where $N$, $U$, and $T$ are small numbers defined later in the paper. The method works as follows: the user records a few training examples of the keyword, and a beam search is used to estimate the labels of the keyword. The algorithm maintains an $N$-best list of label sequence hypotheses to minimize the error that may be incurred by incorrectly estimating the labels. At inference time, each hypothesis is scored using the forward algorithm, and the hypothesis scores are aggregated to obtain a single detection score. 

In the rest of the paper, we describe the proposed method and show that it achieves good performance compared with other query-by-example methods, yet generates easily interpretable models and matches the user's pronunciation better than when the label sequence is supplied through text.

\section{Proposed method}
\label{sec:illust}
\begin{figure}
\underline{\smash{Learning}}
\begin{center}
\begin{tikzpicture}[node distance=2.5cm,auto,>=latex']
    \node [int] (a) {$\phi$};
    \node (b) [left of=a,node distance=2cm, coordinate] {a};
    \node (wav) [left of=b,node distance=0.75cm] {\includegraphics[width=.1\textwidth]{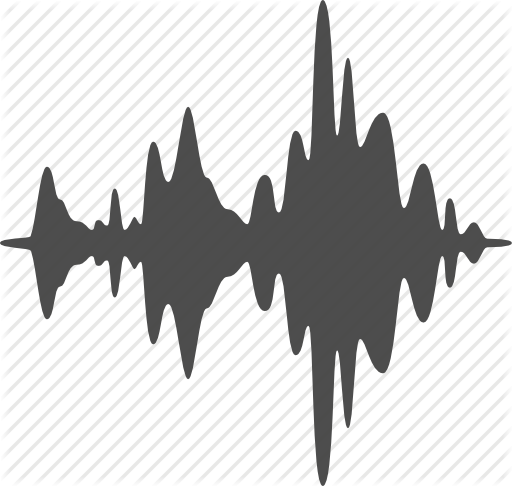}};
    \node [int] (c) [right of=a,node distance=2.5cm,align=center] {Beam\\ Search};
    \path[->] (b) edge node {$\mathbf{x}^{train, 1}$} (a);
    \path[->] (a) edge node {$\bm{\pi}^{train, 1}$} (c);

    \coordinate (arrow_end_1) at (4,0.5);
    \coordinate (arrow_end_2) at (4,0);
    \coordinate (arrow_end_3) at (4,-0.5);
    \draw [->] (c) -- (arrow_end_1);
    \draw [->] (c) -- (arrow_end_2);
    \draw [->, dotted] (c) -- (arrow_end_3);
    \node[text width = 4cm] at ($(arrow_end_1) + (2.2,0)$) {$\mathbf{\hat{y}}^{train, 1}_1$ = \texttt{\{H EH L OW\}}};
    \node[text width = 4cm] at ($(arrow_end_2) + (2.2,0)$) {$\mathbf{\hat{y}}^{train, 1}_2$ = \texttt{\{H AH L OW\}}};
    \node[text width = 4cm] at ($(arrow_end_3) + (2.2,0)$) {\color{gray}\texttt{\{H EH L UW\}}\color{black}};
\end{tikzpicture}

\begin{tikzpicture}[node distance=2.5cm,auto,>=latex']
    \node [int] (a) {$\phi$};
    \node (b) [left of=a,node distance=2cm, coordinate] {a};
    \node (wav) [left of=b,node distance=0.75cm] {\includegraphics[width=.1\textwidth]{speech.png}};
    \node [int] (c) [right of=a,node distance=2.5cm,align=center] {Beam\\ Search};
    \path[->] (b) edge node {$\mathbf{x}^{train, 2}$} (a);
    \path[->] (a) edge node {$\bm{\pi}^{train, 2}$} (c);

    \coordinate (arrow_end_1) at (4,0.5);
    \coordinate (arrow_end_2) at (4,0);
    \coordinate (arrow_end_3) at (4,-0.5);
    \draw [->] (c) -- (arrow_end_1);
    \draw [->] (c) -- (arrow_end_2);
    \draw [->, dotted] (c) -- (arrow_end_3);
    \node[text width = 4cm] at ($(arrow_end_1) + (2.2,0)$) {$\mathbf{\hat{y}}^{train, 2}_1$ = \texttt{\{EH L OW\}}};
    \node[text width = 4cm] at ($(arrow_end_2) + (2.2,0)$) {$\mathbf{\hat{y}}^{train, 2}_2$ = \texttt{\{H EH L OW\}}};
    \node[text width = 4cm] at ($(arrow_end_3) + (2.2,0)$) {\color{gray}\texttt{\{H AH L OW\}}\color{black}};
\end{tikzpicture}
\end{center}

\underline{Inference}
\begin{center}    
\begin{tikzpicture}[node distance=2.5cm,auto,>=latex']
    \node [int] (a) {$\phi$};
    \node (b) [left of=a,node distance=2cm, coordinate] {a};
    \node (wav) [left of=b,node distance=0.75cm] {\includegraphics[width=.1\textwidth]{speech.png}};
    \coordinate (c) at (1.5,0);
    \path[->] (b) edge node {$\mathbf{x}^{test}$} (a);
    \path[-] (a) edge node {$\bm{\pi}^{test}$} (c);

    \coordinate (arrow_end_1) at (2.5,1);
    \coordinate (arrow_end_2) at (2.5,0.3333);
    \coordinate (arrow_end_3) at (2.5,-0.3333);
    \coordinate (arrow_end_4) at (2.5,-1);
    \draw [->] (c) -- (arrow_end_1);
    \draw [->] (c) -- (arrow_end_2);
    \draw [->] (c) -- (arrow_end_3);
    \draw [->] (c) -- (arrow_end_4);
    \node[int, align=center, text width = 4cm] at ($(arrow_end_1) + (2.2,0)$) {$\log  p_{\phi}($\texttt{\{H EH L OW\}}$|\mathbf{x}^{test})$};
    \node[int, align=center, text width = 4cm] at ($(arrow_end_2) + (2.2,0)$) {$\log  p_{\phi}($\texttt{\{H AH L OW\}}$|\mathbf{x}^{test})$};
    \node[int, align=center, text width = 4cm] at ($(arrow_end_3) + (2.2,0)$) {$\log  p_{\phi}($\texttt{\{EH L OW\}}$|\mathbf{x}^{test})$};
    \node[int, align=center, text width = 4cm] at ($(arrow_end_4) + (2.2,0)$) {$\log  p_{\phi}($\texttt{\{H EH L OW\}}$|\mathbf{x}^{test})$};
    
    \coordinate (first) at (6.83,1);
    \coordinate (second) at (6.83,0.3333);
    \coordinate (third) at (6.83,-0.3333);
    \coordinate (fourth) at (6.83,-1);
    \coordinate (join) at (7.83,0);
    \coordinate (end) at (8.33,0);
    \node [text width = 2cm, right of=end,node distance=1.1cm] (score) {$\texttt{score:-3.1}$};
    \draw [-] (first) -- (join);
    \draw [-] (second) -- (join);
    \draw [-] (third) -- (join);
    \draw [-] (fourth) -- (join);
    \draw [->] (join) -- (end);
    
\end{tikzpicture}
    \caption{Illustration of the proposed method. Here, two training examples are recorded by the user, with a beam search of width $B = 3$ and $N = 2$ hypotheses kept per training example.}
    \label{fig:model}
\end{center}
\end{figure}

This section describes the model, learning, and inference for DONUT (Fig. \ref{fig:model}), as well as the memory, storage, and computational requirements.

\subsection{Model}
\label{sec:model}
The proposed method uses a model composed of a wakeword model and a label model. Here we give more detail on these two components.

\subsubsection{Wakeword model}
\label{sec:wakeword_model}

We can model the user's chosen wakeword as a sequence of labels $\mathbf{y} = \{y_u \in \mathcal{A} \mid u = 1,\dots,U\}$, where $\mathcal{A}$ is the set of possible labels, and $U$ is the length of the sequence. The labels could be phonemes, graphemes, or other linguistic subunits; in this work, we use phonemes. It is generally not possible to perfectly estimate $\mathbf{y}$ from only a few training examples. Therefore, we maintain multiple hypotheses as to what the true sequence might be, along with a confidence for each hypothesis, and make use of all of these hypotheses during inference. A trained wakeword model thus consists of a set of label sequences and confidences.

\subsubsection{Label model}
\label{sec:phone_model}
The label model $\phi$ is a neural network trained using CTC on a speech corpus where each audio has a transcript of labels from the label set $\mathcal{A}$. The network accepts an audio in the form of a sequence of acoustic feature vectors $\mathbf{x} = \{x_t \in \mathbb{R}^d \mid t = 1,\dots,T\}$, where $d$ is the number of features per frame, and $T$ is the number of frames. The network outputs a posteriorgram $\bm{\pi} = f_{\phi}(\mathbf{x}) = \{\pi_t \in \mathbb{R}^{1 + |\mathcal{A}|} \mid t = 1,\dots,T\}$ representing the posterior probabilities of each of the labels and the CTC ``blank'' symbol at each timestep.

\subsection{Learning}
\label{sec:learning}
Algorithm \ref{alg:learning} describes the learning phase. The user records three examples of the wakephrase, here denoted by $\mathbf{x}^{train, 1}$, $\mathbf{x}^{train, 2}$, and $\mathbf{x}^{train, 3}$. Once the user has recorded the audios, the label posteriors $\bm{\pi}^{train, i}$ for each audio are computed using the label model $\phi$. The CTCBeamSearch function then runs a beam search of width $B$ over the label posteriors and returns a list of $B$ probable label sequences and their corresponding log probabilities. More details on the beam search algorithm for CTC models can be found in \cite{hannunCTC}. The top $N$ hypotheses $\mathbf{\hat{y}}^{train, i}_{1,\dots,N}$ are kept, and their log probabilities are converted to ``confidences'', which are also stored. Since not every hypothesis is equally good, the confidences can be used to weight the hypotheses during inference. We use an ``acoustic-only'' approach, in the sense that we do not use any sort of language model or pronunciation dictionary to prune the $N$-best list.

\begin{algorithm}
    \caption{Learning}
    \label{alg:learning}
\begin{algorithmic}[1]
\REQUIRE $\mathbf{x}^{train,1}$, $\mathbf{x}^{train, 2}$, $\mathbf{x}^{train, 3}$, $\phi$
\STATE wake\textunderscore model := $\emptyset$
\FOR{$i$ in $(1,2,3)$}
\STATE $\bm{\pi}^{train, i}$ := $f_{\phi}(\mathbf{x}^{train, i})$
\STATE beam, beam\textunderscore scores := CTCBeamSearch($\bm{\pi}^{train, i}$, $B$)
\FOR{$j=1$ to $N$}
    \STATE $\mathbf{\hat{y}}^{train, i}_j$ := beam($j$)
    \STATE $\log p_{\phi}( \mathbf{\hat{y}}^{train,i}_{j} | \mathbf{x}^{train, i})$ := beam\textunderscore scores($j$)
    \STATE $w^{train, i}_j$ := $-\frac{1}{\log p_{\phi}( \mathbf{\hat{y}}^{train,i}_{j} | \mathbf{x}^{train, i})}$
    \STATE wake\textunderscore model := wake\textunderscore model $\cup$  $ (\mathbf{\hat{y}}^{train, i}_j, w^{train, i}_j)$
\ENDFOR
\ENDFOR
\RETURN wake\textunderscore model
\end{algorithmic}
\end{algorithm}

\subsection{Inference}
\label{sec:inference}

Algorithm \ref{alg:inference} describes how the wakeword is detected after the wakeword model has been learned. A voice activity detector (VAD) is used to determine which frames contain speech audio; only these frames are sent to the label model. The VAD thus reduces power consumption by reducing the amount of computation performed by the label model. After the label posteriors are computed by the network, the log probability of each hypothesis in the wakeword model is computed. The CTCForward function returns the log probability of a hypothetical label sequence given the audio by efficiently summing over all possible alignments of the label sequence to the audio \cite{Graves2006}. The log probabilities are weighted by their respective confidences before they are summed to obtain a score. If the score is above a certain pre-determined threshold, the wakeword is detected.

For clarity, we have written Algorithm \ref{alg:inference} as though the posteriors are only computed after a complete audio $\mathbf{x}^{test}$ has been acquired; it is preferable to reduce latency by computing the posteriors and updating the hidden states as each speech frame becomes available from the VAD.
Likewise, the forward algorithm can ingest a slice of $\bm{\pi}^{test}$ at each timestep to compute that timestep's forward probabilities.

\begin{algorithm}
    \caption{Inference}
    \label{alg:inference}
\begin{algorithmic}[1]
\REQUIRE $\mathbf{x}^{test}$, wake\textunderscore model, $\phi$
\STATE $\bm{\pi}^{test}$ := $f_{\phi}(\mathbf{x}^{test})$
\STATE score := 0
\FOR{($\mathbf{\hat{y}}, w$) in wake\textunderscore model}
    \STATE $\log p_{\phi}( \mathbf{\hat{y}}  | \mathbf{x}^{test})$ := CTCForward($\bm{\pi}^{test}$, $\mathbf{\hat{y}}$)
    \STATE score := score + $\log p_{\phi}( \mathbf{\hat{y}}  | \mathbf{x}^{test}) \cdot w$
\ENDFOR
\RETURN score
\end{algorithmic}
\end{algorithm}

\subsection{Runtime requirements}
\label{sec:complexity}
DONUT is fast and suitable for running online on an embedded device. The memory, storage, and computational requirements of running DONUT online can be broken down into two parts: running the label model and running the wakeword model.

The runtime requirements are dominated by the label model (the neural network). The complexity of running the neural network is $O(nT)$, where $n$ is the number of parameters and $T$ is the duration of the audio in frames. We use an RNN with frame stacking \cite{Sak2015}: that is, pairs of contiguous acoustic frames are stacked together so that the RNN operates at 50 Hz instead of 100 Hz, cutting the number of operations in half at the expense of slightly more input-hidden parameters in the first layer.

The wakeword model requires little storage, as it consists of just $3N$ short strings and one real-valued confidence for each string. The CTC forward algorithm requires $O(UT)$ operations to process a single label sequence. If the algorithm is run separately for $N$ hypotheses, and the hypotheses have length $U$ on average, then $O(NUT)$ operations are required. The number of operations could be reduced by identifying and avoiding recomputing shared terms for the forward probabilities (e.g. using a lattice \cite{Chen2016}), at the cost of a more complicated implementation. However, since $N$ and $U$ are small values, this kind of optimization is not crucial. (In the experiments described below, $n$ is 168k, $N$ is 10, and $U$ is on average 10, so it is apparent that in general $O(nT)$ >> $O(NUT)$.) The system requires $O(NU)$ memory to store the forward probabilities for a single timestep; the memory for the previous timestep can be overwritten with the current timestep after the current forward probabilities have been computed.

\section{Experiments}
\label{sec:experiments}

\subsection{Data}
\label{sec:dataset}
All audio data in our experiments is sampled at 16,000 Hz and converted to sequences of 41-dimensional Mel filterbank (FBANK) feature vectors using a 25 ms window with a stride of 10 ms. Here, we describe the two types of datasets used in our experiments: the dataset used to train the label models, and the datasets used to train and test the wakeword detectors.

\paragraph{Label dataset}
We used LibriSpeech \cite{Librispeech}, an English large vocabulary continuous speech recognition (LVCSR) dataset, to train label models. We used the Montreal Forced Aligner \cite{mcauliffe2017montreal} to obtain phoneme-level transcripts written in ARPAbet of the 100- and 360-hour subsets of the dataset. We trained a unidirectional GRU network with 3 layers and 96 hidden units per layer (168k parameters) on LibriSpeech with CTC using the phoneme-level transcripts.

\begin{figure}
    \centering
    \includegraphics[scale=0.38]{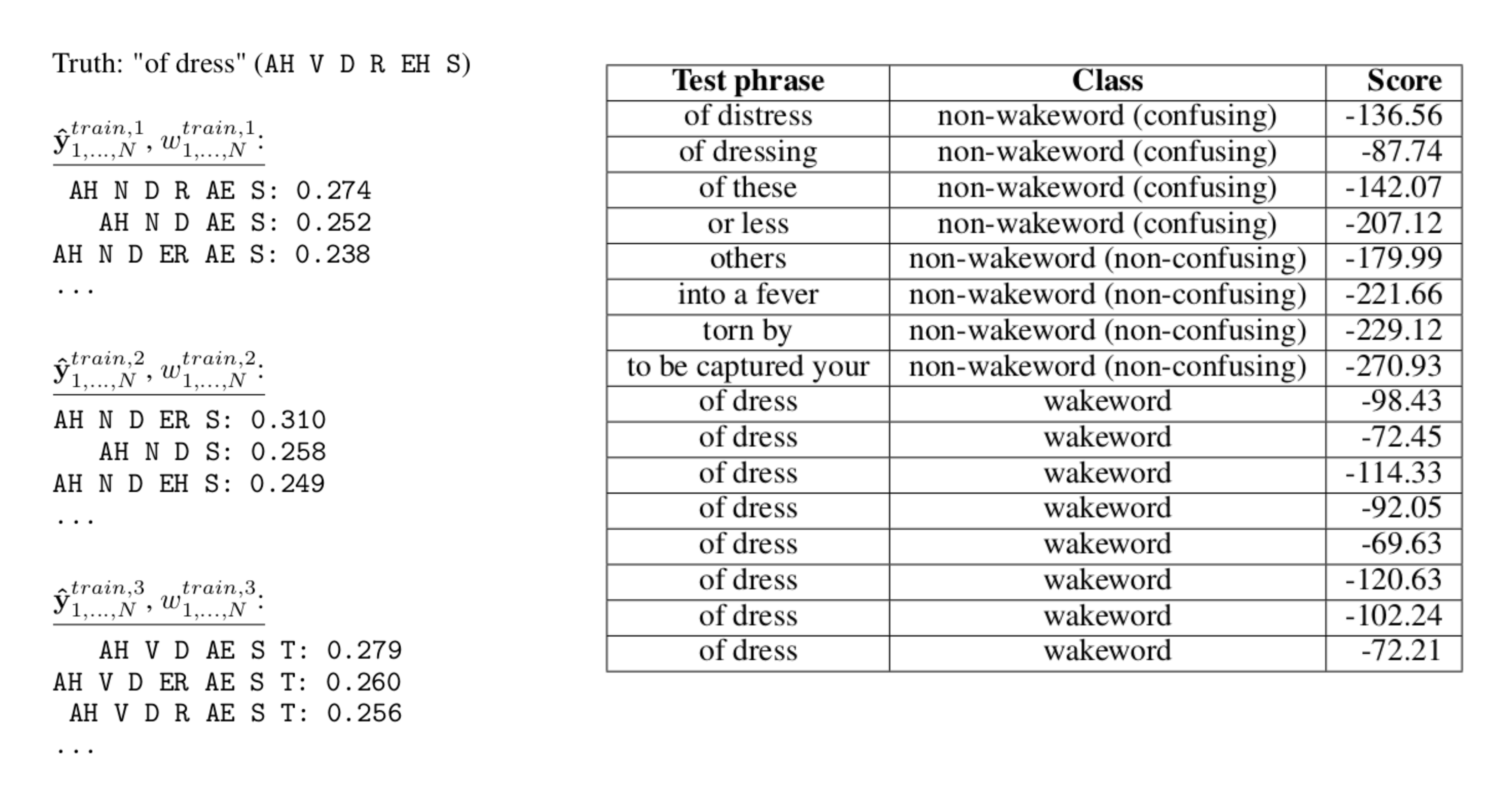}
    
    \caption{Example of a wakeword model generated from three examples of the target ``of dress'' (left) and an episode for that target (right). All examples in this episode are classified correctly given a detection threshold of -121, except for ``of dressing'', which contains the wakeword as a substring.}
    \label{fig:episode}
\end{figure}

\paragraph{Wakeword datasets}
We created two wakeword datasets: one based on the 500-hour subset of LibriSpeech (LibriSpeech-Fewshot) and one based on crowdsourced English recordings (English-Fewshot). Both datasets are composed of a number of few-shot learning ``episodes''. Each episode contains support examples and test examples.  The support set contains three examples of the target phrase spoken by a single speaker. The test set contains a number of positive and negative examples. An example of an episode is shown in Fig. \ref{fig:episode}. The episodes are split into one subset for hyperparameter tuning and another subset for reporting performance.

To create the LibriSpeech-Fewshot dataset, we split the LibriSpeech recordings into short phrases between 500 ms and 1,500 ms long, containing between one and four words.  These short phrases were selected and grouped together to form 6,047 episodes. The test set contains eight positive examples by the same speaker and 24 negative examples by random speakers.  Of the negative examples, twenty are phonetically similar (``confusing''), and four are phonetically dissimilar (``non-confusing''). To produce the confusing examples, we generated a phoneme-level transcript for each example, calculated the phoneme edit distance between the target phrase and all other available phrases, and chose the 20 phrases with the lowest phoneme edit distance. The non-confusing examples were chosen at random from the remaining phrases. 

To create the English-Fewshot dataset, we used crowdsourcing to record speakers saying phrases consisting of ``Hello'' followed by another word: for example, ``Hello Computer''. Like the LibriSpeech-Fewshot dataset, this dataset has positive examples from the same speaker and negative examples from different speakers; however, here there are also negative examples from the same speaker, so as to show that the models are not simply performing speaker verification. Due to data-gathering constraints, we were unable to obtain ``imposter'' examples in which a different speaker says the target phrase, but we plan to explore this in the future.

All wakeword models used beam width $B=100$ and kept $N = 10$ hypotheses per training example. We use the receiver operating characteristic (ROC) curve to measure the performance of a wakeword detector. A single detection threshold is used across all episodes. Two performance metrics are reported: the equal error rate (EER; lower is better) and the area-under-ROC-curve (AUC; higher is better) metric. An EER of 0\% or an AUC of 1 indicates a perfect classifier. 

\subsection{Comparison with other query-by-example methods}
In the first experiment, we compare the performance of DONUT with two other query-by-example keyword spotting methods: dynamic time warping (DTW) based on the raw FBANK input and DTW based on the posteriorgram (the output of the label model). We used the $\ell_2$ norm to compare FBANK features, and we used the distance-like metric suggested in \cite{PosteriorgramDTW} to compare posteriorgram features:
\begin{equation}
    d(p,q) = -\log((\lambda u + (1-\lambda) p) \cdot (\lambda u + (1-\lambda) q)),
\end{equation}
where $\lambda$ is a small positive number (we used $1\mathrm{e}{-5}$) and $u$ is a uniform distribution (a vector with entries equal to $\frac{1}{1+|\mathcal{A}|}$; used to prevent $\log(0)$ by smoothing the peaky output distribution). We also tried removing the softmax, using the $\ell_2$ norm as the distance metric, and using a label model trained using the framewise cross-entropy loss instead of the CTC loss. None of these modifications improved performance; we report the best result with the CTC model here. 

Table \ref{QbE_table} shows the performance of the query-by-example methods on English-Fewshot. We report the performance for three separate cases, in decreasing order of difficulty: the cases when the negative examples are 1) confusing and taken from the same speaker, 2) non-confusing and taken from the same speaker, and 3) non-confusing and taken from different speakers. DONUT outperforms both DTW methods in all three cases.

\begin{table}
  \caption{Performance on English-Fewshot of query-by-example methods.}
  \label{QbE_table}
  \centering
  \begin{tabular}{lcccccc}
    \toprule
           & \multicolumn{2}{c}{Same speaker} & \multicolumn{2}{c}{Same speaker} & \multicolumn{2}{c}{Different speaker}   \\
                  & \multicolumn{2}{c}{(confusing)} & \multicolumn{2}{c}{(non-confusing)} & \multicolumn{2}{c}{(non-confusing)}                   \\
    \cmidrule(lr){2-3} \cmidrule(lr){4-5} \cmidrule(lr){6-7}
    Method     & EER & AUC & EER & AUC & EER & AUC \\
    \midrule
    DTW (FBANK)  & 24.2\% & 0.839  & 32.4\%& 0.745  & 19.3\% & 0.893\\
    DTW (posteriorgram) & 20.8\% & 0.877 & 17.3\% & 0.912 & 11.0\%  &  0.956   \\
    DONUT  & \textbf{7.8\%} & \textbf{0.975} & \textbf{7.3\%} & \textbf{0.977} & \textbf{3.7\%} & \textbf{0.993} \\
    \bottomrule
  \end{tabular}
\end{table}

\subsection{Comparison with query-by-string method}
In this experiment, we compare the performance of our method with the performance of conventional CTC keyword spotting when the ``true'' label sequence is provided (e.g., by the user through a text interface). The phoneme sequence for each phrase in the LibriSpeech-Fewshot dataset was obtained using forced alignment and used as the wakeword model for each episode.

Table \ref{table:oracle-table} shows that for phonetically confusing examples, DONUT outperforms the text-based approach, and for non-confusing examples, the two approaches perform roughly the same, with the text-based approach performing very slightly better. This result indicates that not only does DONUT provide a more convenient interface than query-by-string keyword spotting, it also has the same or even better performance.

\begin{table}
  \caption{Performance on LibriSpeech-Fewshot compared to query-by-string CTC.}
  \label{table:oracle-table}
  \centering
  \begin{tabular}{lcccc}
    \toprule
      & \multicolumn{2}{c}{Confusing} & \multicolumn{2}{c}{Non-confusing}                    \\
    \cmidrule(lr){2-3} \cmidrule(lr){4-5}
    Method   & EER & AUC & EER & AUC \\
    \midrule
    CTC (query-by-string) & 26.9\% & 0.810 & \textbf{9.6\%} & \textbf{0.968} \\
    DONUT   & \textbf{21.0\%} & \textbf{0.872} & 9.6\% & 0.966 \\
    \bottomrule
  \end{tabular}
\end{table}

\subsection{Interpretability of the model}
Like conventional CTC keyword spotting, DONUT is interpretable, which makes it easy for a system designer to identify problems with the model and improve it. For example, Fig. \ref{fig:episode} shows an example of a wakeword model learned for the phrase ``of dress''. In the first two training examples, the network hears an ``\texttt{N}'' sound where one would expect the ``\texttt{V}'' phoneme in ``of''. This information can be used to improve the model: one could retrain the label model with more examples short words such as ``of'' and ``on'', to help the model distinguish short sounds more easily. Alternately, it could become apparent after listening to the training examples that for the speaker's personal accent the phrase does indeed contain an  ``\texttt{N}'' sound. 

Debugging the inference phase is also made easier by the use of CTC. It is possible to decode phoneme sequences from the test audio using a beam search, although this is not necessary to do during inference. One could inspect the decoded sequences from an audio that causes a false accept to identify hypotheses that should be removed from the model to make the false accept less likely to occur. If a false reject occurs, one could check whether the wakeword model hypotheses are found in the decoded sequences or if the network hears something completely different.

\subsection{Impact of hyperparameters on performance}
DONUT has a few hyperparameters: the beam width $B$, the number of hypotheses kept from the beam search $N$, the label model $\phi$, and the way in which the hypothesis scores are aggregated. Here we explore the impact of these hyperparameters on performance using the English-Fewshot dataset.

Increasing the number of hypotheses generally improves performance (Table \ref{num_hypotheses}), though we have found that this may yield diminishing returns. Even a simple greedy search ($B=1$, $N=1$), which can be implemented by picking the top output at each timestep, works fairly well for our system.

\begin{table}
  \caption{Impact of number of hypotheses on performance.}
  \label{num_hypotheses}
  \centering
  \begin{tabular}{lll}
    \toprule
    Beam width $B$ & \# of kept hypotheses $N$ & EER \\
    \midrule
    1 (greedy) & 1  & 4.2\%  \\
    100 & 1 & 4.4\%  \\
    100 & 2 & 4.2\%  \\
    100 & 5 & 3.8\%  \\
    100 & 10 & 3.7\%  \\
    100 & 20 & 3.4\%  \\
    100 & 50 & 3.3\%  \\
    100 & 100 & 3.1\%  \\
    \bottomrule
  \end{tabular}
\end{table}
With respect to the impact of the choice of label model, we find that label models with lower phoneme error rate (edit distance between the true label sequence and the model's prediction) for the original corpus they were trained on have a lower error rate for wakeword detection (Table \ref{num_parameters}). This suggests that making an improvement to the label model can be expected to translate directly to a decrease in EER/increase in AUC.

\begin{table}
  \caption{Impact of label model quality on performance.}
  \label{num_parameters}
  \centering
  \begin{tabular}{lll}
    \toprule
    Model size & Phoneme error rate & EER \\
    \midrule
    2x128 (186k params) & 17.5\%  &  5.7\%    \\
    3x96 (168k params) & 15.8\%  &  3.7\%  \\
    3x512 (4m params) & \textbf{11.5\%} &   \textbf{1.4\%} \\
    \bottomrule
  \end{tabular}
\end{table}

In the inference algorithm described above (Algorithm \ref{alg:inference}), the hypotheses' scores are aggregated by taking a weighted sum, where each weight is the inverse of the log probability of that hypothesis given its corresponding training example. Without the weighting, performance was hurt because some hypotheses are a worse fit to the data than others. 
A more principled approach to aggregating the scores might be to treat the hypotheses' log probabilities from training as log priors and add them to the scores, since multiplying by a prior is equivalent to adding a log prior, and to take the $\texttt{logsumexp}()$ of the scores plus their log priors, since adding two probabilities is equivalent to taking the $\texttt{logsumexp}()$ of two log probabilities.
However, we have found that this does not work as well as the weighted sum approach, perhaps because the $\texttt{logsumexp}()$ function acts like $\texttt{max}()$ and tends to pick out a single hypothesis instead of smoothly blending the hypotheses. 

\section{Conclusion}
\label{sec:ref}
In this paper, we proposed DONUT, an efficient algorithm for online query-by-example keyword spotting using CTC. The algorithm learns a list of hypothetical label sequences from the user's speech during enrollment and uses these hypotheses to score audios at test time. We showed that the model is interpretable, and thus easy to inspect, debug, and tweak, yet at the same time has high accuracy. Because training a wakeword model amounts to a simple beam search, it is possible to train a model on the user's device without uploading a user's private voice data to the cloud.

Our technique is in principle applicable to any domain in which a user would like to teach a system to recognize a sequence of events, such as a melody (a sequence of musical notes) or a gesture (a sequence of hand movements). It would be interesting to see how well the proposed technique transfers to these other domains.

\bibliographystyle{IEEEbib}
\bibliography{refs}

\begin{thebibliography}{10}

\bibitem{Chen2014}
Guoguo Chen, Carolina Parada, and Georg Heigold,
\newblock ``{Small-footprint keyword spotting using deep neural networks},''
\newblock {\em ICASSP}, 2014.

\bibitem{Myer}
Samuel Myer and Vikrant~Singh Tomar,
\newblock ``{Efficient keyword spotting using time delay neural networks},''
\newblock {\em Interspeech}, 2018.

\bibitem{Graves2006}
Alex Graves, Santiago Fern{\'a}ndez, Faustino Gomez, and J{\"u}rgen
  Schmidhuber,
\newblock ``{Connectionist Temporal Classification: Labelling Unsegmented
  Sequence Data with Recurrent Neural Networks},''
\newblock {\em ICML}, 2006.

\bibitem{hwang2015online}
Kyuyeon Hwang, Minjae Lee, and Wonyong Sung,
\newblock ``Online keyword spotting with a character-level recurrent neural
  network,''
\newblock {\em arXiv preprint arXiv:1512.08903}, 2015.

\bibitem{Lengerich2016}
Chris Lengerich and Awni Hannun,
\newblock ``An end-to-end architecture for keyword spotting and voice activity
  detection,''
\newblock {\em NIPS}, 2016.

\bibitem{Zhuang2016}
Yimeng Zhuang, Xuankai Chang, Yanmin Qian, and Kai Yu,
\newblock ``{Unrestricted vocabulary keyword spotting using LSTM-CTC},''
\newblock {\em Interspeech}, 2016.

\bibitem{snips}
Thibault Gisselbrecht,
\newblock ``{Machine Learning on Voice: a gentle introduction with Snips
  Personal Wake Word Detector},''
\newblock 2018,
\newblock
  https://medium.com/snips-ai/machine-learning-on-voice-a-gentle-introduction-with-snips-personal-wake-word-detector-133bd6fb568e.

\bibitem{PosteriorgramDTW}
Timothy~J Hazen, Wade Shen, and Christopher White,
\newblock ``Query-by-example spoken term detection using phonetic posteriorgram
  templates,''
\newblock {\em ASRU}, 2009.

\bibitem{Zhang2009}
Yaodong Zhang and James~R. Glass,
\newblock ``{Unsupervised spoken keyword spotting via segmental DTW on Gaussian
  posteriorgrams},''
\newblock {\em ASRU}, 2009.

\bibitem{Rodriguez-Fuentes2014}
Luis~J. Rodriguez-Fuentes, Amparo Varona, Mikel Penagarikano, Germ{\'{a}}n
  Bordel, and Mireia Diez,
\newblock ``{High-performance Query-by-Example Spoken Term Detection on the SWS
  2013 evaluation},''
\newblock {\em ICASSP}, 2014.

\bibitem{Chen2015}
Guoguo Chen, Carolina Parada, and Tara~N. Sainath,
\newblock ``{Query-by-example keyword spotting using Long Short Term Memory
  Networks},''
\newblock {\em ICASSP}, 2015.

\bibitem{Settle2016}
Shane Settle and Karen Livescu,
\newblock ``{Discriminative Acoustic Word Embeddings: Recurrent Neural
  Network-Based Approaches},''
\newblock {\em SLT}, 2016.

\bibitem{Settle2017}
Shane Settle, Keith Levin, Herman Kamper, and Karen Livescu,
\newblock ``Query-by-example search with discriminative neural acoustic word
  embeddings,''
\newblock {\em Interspeech}, 2017.

\bibitem{radford2017learning}
Alec Radford, Rafal Jozefowicz, and Ilya Sutskever,
\newblock ``Learning to generate reviews and discovering sentiment,''
\newblock {\em arXiv preprint arXiv:1704.01444}, 2017.

\bibitem{verwimp2018}
Lyan Verwimp, Hugo Van~hamme, Vincent Renkens, and Patrick Wambacq,
\newblock ``{State Gradients for RNN Memory Analysis},''
\newblock {\em Interspeech}, 2018.

\bibitem{bluche2015framewise}
Th{\'e}odore Bluche, Hermann Ney, J{\'e}r{\^o}me Louradour, and Christopher
  Kermorvant,
\newblock ``Framewise and {CTC} training of neural networks for handwriting
  recognition,''
\newblock {\em ICDAR}, 2015.

\bibitem{hannunCTC}
Awni Hannun,
\newblock ``Sequence modeling with {CTC},''
\newblock {\em Distill}, 2017,
\newblock https://distill.pub/2017/ctc.

\bibitem{Sak2015}
Haşim Sak, Andrew Senior, Kanishka Rao, and Fran{\c{c}}oise Beaufays,
\newblock ``{Fast and Accurate Recurrent Neural Network Acoustic Models for
  Speech Recognition},''
\newblock {\em Interspeech}, 2015.

\bibitem{Chen2016}
Zhehuai Chen, Wei Deng, Tao Xu, and Kai Yu,
\newblock ``{Phone synchronous decoding with CTC lattice},''
\newblock {\em Interspeech}, 2016.

\bibitem{Librispeech}
Vassil Panayotov, Guoguo Chen, Daniel Povey, and Sanjeev Khudanpur,
\newblock ``{LibriSpeech}: an {ASR} corpus based on public domain audio
  books,''
\newblock {\em ICASSP}, 2015.

\bibitem{mcauliffe2017montreal}
Michael McAuliffe, Michaela Socolof, Sarah Mihuc, Michael Wagner, and Morgan
  Sonderegger,
\newblock ``{Montreal Forced Aligner: Trainable text-speech alignment using
  Kaldi},''
\newblock {\em Interspeech}, 2017.

\end{thebibliography}

\end{document}